\documentclass[conference]{IEEEtran}

\IEEEoverridecommandlockouts
\usepackage{cite}
\usepackage{amsmath,amssymb,amsfonts}
\usepackage{algorithmic}
\usepackage{graphicx}
\usepackage{textcomp}
\usepackage[table]{xcolor}
\def\BibTeX{{\rm B\kern-.05em{\sc i\kern-.025em b}\kern-.08em
    T\kern-.1667em\lower.7ex\hbox{E}\kern-.125emX}}

\usepackage[hidelinks]{hyperref}
\hypersetup{
   linkcolor=blue,
   breaklinks=true,   
   colorlinks=true,   
   citecolor=blue,
   urlcolor=blue
}

\usepackage{siunitx,booktabs}
\usepackage{multirow}
\usepackage{caption}
\usepackage{hhline}
\usepackage{comment}

\usepackage[ruled,vlined,linesnumbered]{algorithm2e}

\SetCommentSty{mycommfont}
\SetKwComment{Comment}{$\triangleright$\ }{}
\let\oldnl\nl%
\newcommand{\nonl}{\renewcommand{\nl}{\let\nl\oldnl}}%

\newcommand \name{GAIDE\xspace}

\newcommand{\xxnote}[3]{}
\renewcommand{\xxnote}[3]{\color{#2}{#1: #3}}

\begin{document}

\title{\huge{\textbf{GAIDE:} \textbf{\underline{G}}raph-based \textbf{\underline{A}}ttention Masking for Spat\textbf{\underline{i}}al- and Embo\textbf{\underline{d}}im\textbf{\underline{e}}nt-aware Motion Planning}}

\author{Davood Soleymanzadeh$^{1}$, Xiao Liang$^{2,*}$, and Minghui Zheng$^{1,*}$
\thanks{$^{1}$Davood Soleymanzadeh and Minghui Zheng are with the J. Mike Walker '66 Department of Mechanical Engineering, Texas A\&M University, College Station, TX 77843, USA (\tt\footnotesize e-mail: davoodso@tamu.edu; mhzheng@tamu.edu).}
\thanks{$^{2}$Xiao Liang is with the Zachry Department of Civil and Environmental
Engineering, Texas A\&M University, College Station, TX 77843 USA (\tt\footnotesize e-mail:
xliang@tamu.edu).}
\thanks{$^*$ Corresponding Authors.}
\thanks{This work was supported by the USA National Science Foundation under Grant No. 2527316 and No. 2422826. Portions of this research were conducted with the advanced computing resources provided by Texas A\&M High Performance Research Computing.}
}
\maketitle
\begin{abstract}
Sampling-based motion planning algorithms are widely used for motion planning of robotic manipulators, but they often struggle with sample inefficiency in high-dimensional configuration spaces due to their reliance on uniform or hand-crafted informed sampling primitives. Neural informed samplers address this limitation by learning the sampling distribution from prior planning experience to guide the motion planner towards planning goal. However, existing approaches often struggle to encode the spatial structure inherent in motion planning problems. To address this limitation, we introduce Graph-based Attention Masking for Spatial- and Embodiment-aware Motion Planning (GAIDE), a neural informed sampler that leverages both the spatial structure of the planning problem and the robotic manipulator's embodiment to guide the planning algorithm. GAIDE represents these structures as a graph and integrates it into a transformer-based neural sampler through attention masking. We evaluate GAIDE against baseline state-of-the-art sampling-based planners using uniform sampling, hand-crafted informed sampling, and neural informed sampling primitives. Evaluation results demonstrate that GAIDE improves planning efficiency and success rate. A brief video introduction of this work is available via this \href{https://zh.engr.tamu.edu/wp-content/uploads/sites/310/2026/03/GAIDE-DEMO-Small.mp4}{link} and as the supplemental document.  
\end{abstract}

\begin{IEEEkeywords}
Motion Planning, Neural Informed Sampling, Transformers, Attention Mask
\end{IEEEkeywords}
\section{Introduction}
Motion planning for robotic manipulators aims to find a feasible, collision-free path that connects a start and goal configuration in a high-dimensional configuration space \cite{soleymanzadeh2026towards}. A broad class of algorithms has been developed to address this problem \cite{soleymanzadeh2026towards}. Among them, probabilistically complete sampling-based motion planners have demonstrated strong performance for robotic manipulators with high-dimensional configuration spaces due to their scalability \cite{lavalle2006planning}.

Sampling-based motion planning algorithms construct a tree toward the planning goal by iteratively (1) sampling configurations within the configuration space, (2) steering towards these samples, and (3) collision checking such connections \cite{lavalle2006planning}. These planners either rely on uniform sampling \cite{lavalle1998rapidly} or hand-crafted informed samplers for sample generation \cite{strub2020adaptively}. However, uniform sampling is computationally inefficient in high-dimensional spaces, while hand-crafted informed samplers are often sensitive to initialization and difficult to design for high-dimensional configuration spaces \cite{soleymanzadeh2025simpnet}.

\begin{figure}[t] 
\centering
\includegraphics[width=0.39\textwidth]{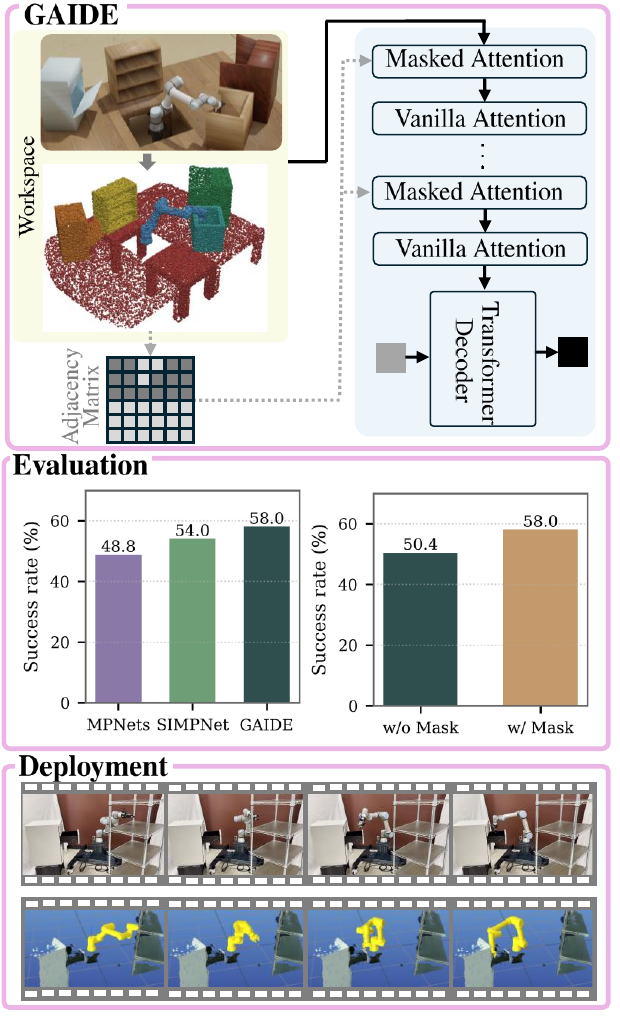}
\caption{\textbf{An overview of \name}: the proposed neural informed sampler constructs a graph that represents both the robotic manipulator embodiment and the spatial relationships inherent within the motion planning problem. The adjacency matrix of this graph is incorporated into a transformer-based neural sampler via attention masking to enable spatial- and embodiment-aware informed sampling. The framework is embedded within sampling-based planners as an informed sampler and demonstrates superior performance compared to benchmark neural samplers, MPNets \cite{qureshi2020motion} and SIMPNet \cite{soleymanzadeh2025simpnet}.}
\label{fig_1- simpt_overview}
\end{figure}

\begin{figure}[htbp] 
\centering
\includegraphics[width=0.4\textwidth]{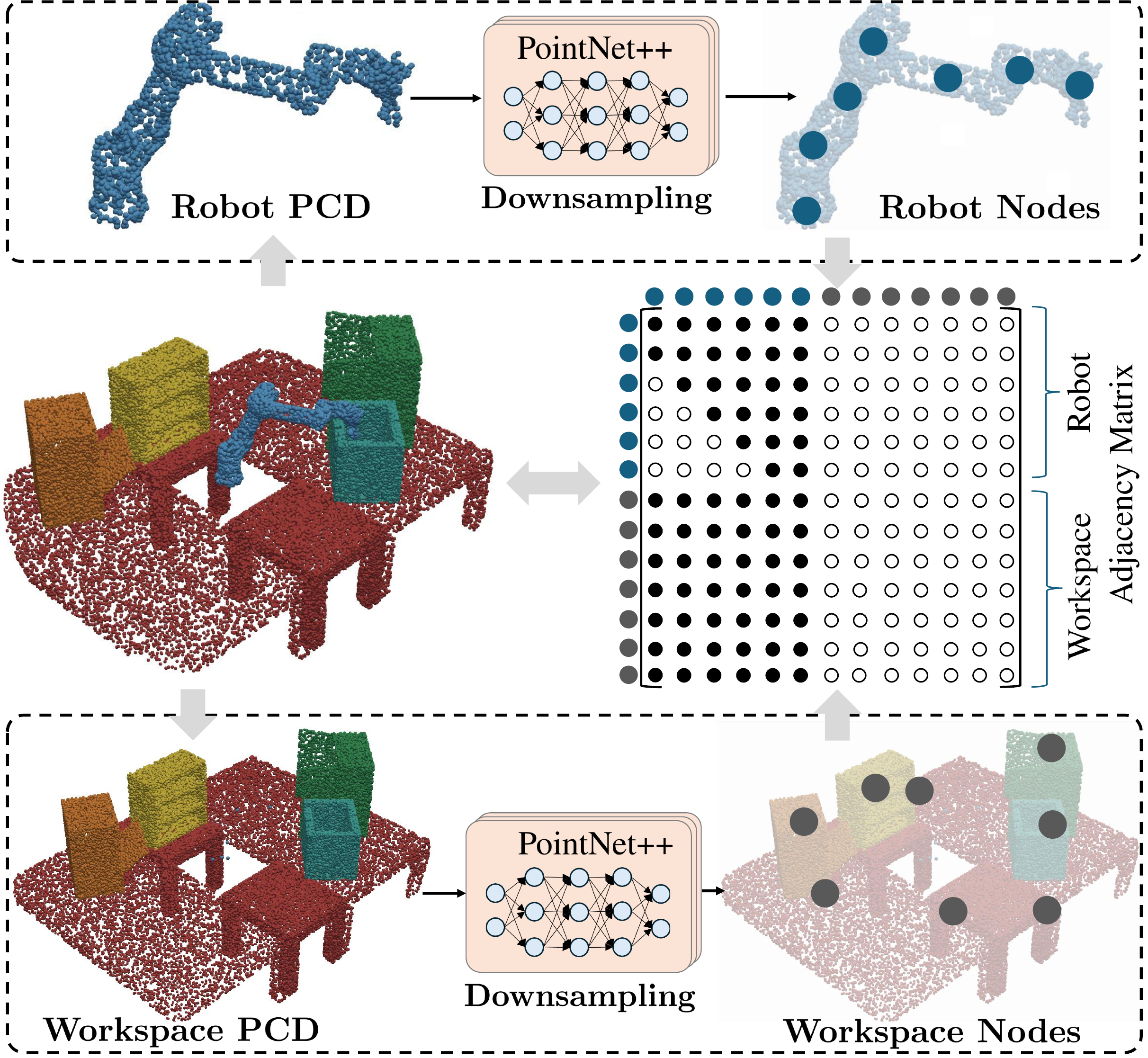}
\caption{\textbf{Graph Construction}: an illustration of spatial and embodiment graph. An undirected graph is constructed over the downsampled manipulator point cloud to implicitly encode the manipulator's kinematics chain, while a directed graph connects the downsampled workspace point cloud to the robot nodes to capture the inherent spatial relationships within the motion planning problem. ``PCD'' denotes point cloud and ``PointNet++'' is set abstraction layers from PointNet++ \cite{qi2017pointnet++}.}
\label{fig_2- graphconstruction}
\end{figure}

Recent advances in neural motion planning aim to improve sampling efficiency by learning informed sampling distributions from planning datasets. These approaches train neural samplers on oracle-generated paths to bias sampling toward promising regions \cite{qureshi2020motion}. Despite their success, most existing neural samplers do not encode the spatial structure of the planning space or the robotic manipulator's embodiment \cite{zang2022robot}. To address this challenge, SIMPNet \cite{soleymanzadeh2025simpnet} introduced an embodiment-aware informed sampler by constructing a graph over the manipulator embodiment, and leveraging message-passing graph neural networks (GNNs) \cite{battaglia2018relational} for informed sampling. However, GNNs often struggle to capture the long-range interactions and dependencies inherent within motion planning due to representation oversmoothing and oversquashing in deep message-passing networks \cite{alon2020bottleneck}.

In this work, we introduce \name, a spatial- and embodiment-aware neural motion planner that encodes both kinematic and planning space structure. \name constructs a unified graph that represents (1) the manipulator's kinematics structure and (2) the spatial relationship between robot and planning environment. Instead of relying on message passing, we integrate this graph into a transformer-based neural sampler by incorporating the graph adjacency matrix as an attention mask \cite{sferrazza2024body}. This structured attention mechanism constrains information flow according to the underlying planning structure while preserving the transformer’s ability to model long-range dependencies. Embedded within a bidirectional sampling-based planner \cite{qureshi2020motion}, \name reduces planning time and enhances success rate compared to state-of-the-art planners using uniform sampling, hand-crafted informed sampling, and neural informed sampling.

The main contributions of this work are:
\begin{itemize}
    \item We construct a graph that captures both the kinematic structure of the manipulator and the spatial structure of the planning scene. This graph is incorporated into a transformer-based neural sampler through structured attention masking, enabling spatial- and embodiment-aware informed sampling. 

    \item We evaluate \name across diverse held-out planning tasks and benchmark it against state-of-the-art sampling-based planners using uniform, heuristic-based, and neural informed sampling strategies, in terms of planning time, path cost, and success rate.
\end{itemize}

This paper is organized as follows: Section \ref{sec: relatedwork} reviews related work in manipulator motion planning. Section \ref{sec: simpt} presents the \name framework. Section \ref{sec: discussions} presents the evaluation results and comparisons with the benchmark planners. Section \ref{sec: conclusions} concludes the paper.
\section{Related Works} \label{sec: relatedwork}
In this section, we review prior work in robotic manipulator motion planning and highlight the key distinctions between \name and existing approaches.

\vspace{0.1cm}
\noindent 
\textbf{Robotic manipulator motion planning.} Decades of research have led to a wide range of motion planning algorithms for robotic manipulators, most notably sampling-based \cite{lavalle1998rapidly, strub2020adaptively} and trajectory optimization algorithms \cite{mukadam2016gaussian}. Sampling-based algorithms construct a tree connecting the start and goal configurations by uniformly or inform sampling the configuration space. Trajectory optimization algorithms subject an initial, often in-collision, path to planning constraints such as collision avoidance and path smoothness. However, the uniform \cite{lavalle1998rapidly} and hand-crafted informed \cite{strub2020adaptively} samplings of sampling-based algorithms are sample inefficient for high DOF robotic manipulators operating in cluttered workspaces \cite{soleymanzadeh2025simpnet}. Trajectory optimization algorithms are highly sensitive to the initial guess and may get stuck in local minima \cite{ichnowski2020deep}. \name addresses these challenges by incorporating the kinematic and spatial structure of the planning problem to remedy the sample inefficiency of sampling-based planners with informed sample generation. In addition, \name spatial-informed initial trajectories can be potentially used to warm-start trajectory optimization algorithms to reduce sensitivity to local minima.

\begin{figure*}[htbp] 
\centering
\includegraphics[width=0.8\textwidth]{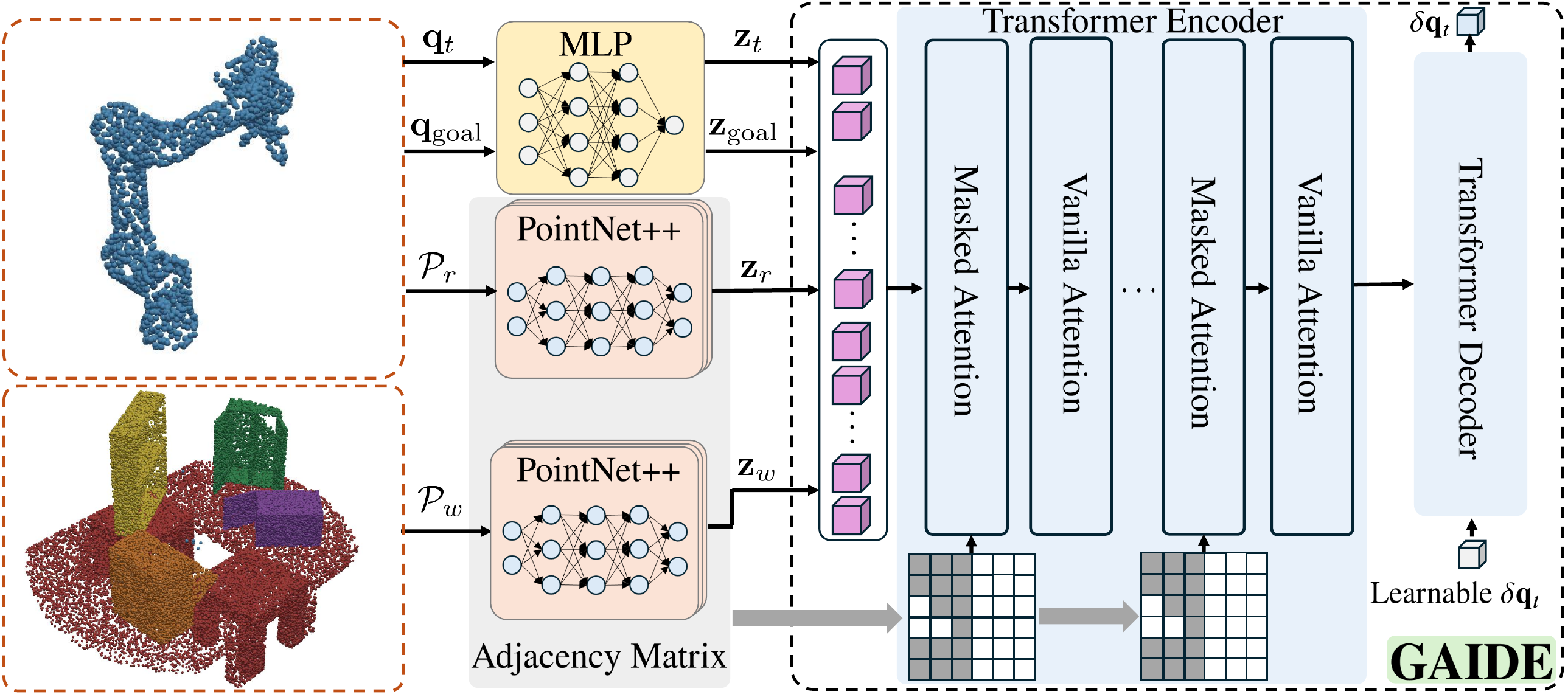}
\caption{\textbf{\name network architecture:} The framework leverages current-time-step workspace information (including the workspace and robot point clouds) together with configuration space features (current-time-step and goal configuration) to generate an informed sample that guides the robotic manipulator toward the motion planning goal. $\mathbf{q}_t,~\mathcal{P}_r$, and $\mathbf{q}_{\text{goal}}$ are the current time-step configuration, current-time step robotic manipulator point cloud, and motion planning goal configuration, respectively. $\mathbf{z}_t$, $\mathbf{z}_{\text{goal}}$, $\mathbf{z}_r$, $\mathbf{z}_w$, and $\delta \mathbf{q}_t$ are the current time-step configuration, planning goal configuration, robot point cloud, scene point cloud embeddings, and predicted joint angle, respectively. ``PointNet++'' denotes set abstraction layers from PoinNet++ \cite{qi2017pointnet++}.}
\label{fig_3- simptstructure}
\end{figure*}

\vspace{0.1cm}
\noindent
\textbf{Neural motion planning for robotic manipulators.} Recent advances in deep learning have enabled neural approaches for motion planning, either as end-to-end planners \cite{dalal2024neural,yang2025deep, soleymanzadeh2025perfact}, or as learned components integrated into classical motion planning algorithms. Within sampling-based planners, neural networks have been used as informed samplers \cite{qureshi2020motion}, and implicit collision checkers \cite{kim2022graphdistnet}, while in trajectory optimization algorithms, they have been employed to encode the planning distribution to warm-start trajectory optimization \cite{carvalho2023motion}. However, most existing approaches struggle to capture the manipulator embodiment and spatial structure inherent within the motion planning problems. \name explicitly models these structures through a graph representation and embeds them directly into the neural sampler via attention masking.

\vspace{0.1cm}
\noindent
\textbf{Spatial-informed neural planning for robotic manipulators.} A growing line of work seeks to explicitly model spatial relationships within planning problems using graph representations. GNNs provide a natural framework for processing structured data and have been applied to informed sampling \cite{soleymanzadeh2025simpnet}, edge evaluation \cite{yu2021reducing}, and collision checking \cite{kim2022graphdistnet}. These methods construct graphs to encode geometric or kinematic structure and rely on message passing to propagate relational information \cite{battaglia2018relational}. However, GNNs often struggle to model long-term interactions and dependencies due to representation oversmoothing and overquashing in deep message passing networks \cite{alon2020bottleneck}. To overcome this limitation, \name encodes the graph constructed based on the manipulator embodiment and planning spatial structure directly into the planning network via attention masking to enable efficient motion planning.

\section{Graph-based Attention Masking for Spatial- and Embodiment-aware Motion Planning (\name)} \label{sec: simpt}
In this section, we introduce \name, a neural informed sampler that encodes both the manipulator embodiment and spatial structure within planning problems to improve the performance of sampling-based motion planning algorithms for high DOF robotic manipulators. Figure \ref{fig_3- simptstructure} demonstrates the structure of \name.

\subsection{Motion Planning Definition}
Let the configuration space of an $n$-DOF robotic manipulator be denoted as $\mathcal{C} \in \mathbb{R}^n$ spanned by its joint values. The obstacle and free configuration spaces are defined as $\mathcal{C}_{obs} \subset \mathcal{C}$ and $\mathcal{C}_{free} = \mathcal{C} \backslash \mathcal{C}_{obs}$, respectively. The goal of the motion planning problem is to find a feasible path $\sigma = [\mathbf{q}_1, \cdots, \mathbf{q}_t, \cdots, \mathbf{q}_T]$ connecting a start configuration ($\mathbf{q}_{\text{start}} \in \mathcal{C}_{free}$) and a goal configuration ($\mathbf{q}_{\text{goal}} \in \mathcal{C}_{free}$) such that:
\begin{equation} \label{ppdef}
\begin{aligned}
    \sigma(0) &= \mathbf{q}_{\text{start}}, \\
    \sigma(t) &\in \mathcal{C}_{\text{free}}, \\
    \sigma(T) &= \mathbf{q}_{\text{goal}}.
\end{aligned}
\end{equation}

\subsection{Planning Embodiment and Spatial Graph Representation}
Let $G = (V, E)$ denote a graph, where $V$ is the set of nodes, and $E$ is the set of edges. An edge $e_{ij} \in E$ connects node $v_i \in V$ and $v_j \in V$ if they are adjacent. The adjacency matrix $A \in \mathbb{R}^{n_v \times n_v}$ is defined such that $A_{ij}=1$ if $e_{ij} \in E$ and $A_{ij}=0$ if $e_{ij} \notin E$ where $n_v$ denotes the number of nodes in the graph.

\begin{algorithm}[t]
\caption{Bidirectional Motion Planning}
\label{alg: simpt}
    \DontPrintSemicolon
    \textbf{Given:} Neural sampler (\name): $\pi_\theta$. \\
    \textbf{Given:} Start and goal configurations: $\mathbf{q}_{\text{start}}$; $\mathbf{q}_{\text{goal}}$. \\
    \textbf{Given:} Scene point cloud, and robot point cloud sampler: \textit{ScenePCD}, \textit{PCDSampler}. \\

    \BlankLine
    $\mathbf{q}^a \leftarrow \{\mathbf{q}_{start}\}$, $\mathbf{q}^b \leftarrow \{\mathbf{q}_{goal}\}$ \\
    $\mathbf{q}\leftarrow \emptyset$ \\
    \textit{Complete} $\leftarrow$ \textit{False} 
    
    \BlankLine
    \For{$i\leftarrow0$ to steps}{
        \Comment*[l]{Sample point cloud}
        \textit{RobotPCD}$_a$ $\leftarrow$ \textit{PCDSampler}($\mathbf{q}^a$(end)) \\
        \Comment*[l]{Neural informed sampling}
        $\delta\mathbf{q}_{t}\leftarrow~\pi_\theta(\mathbf{q}^a(\text{end}), \mathbf{q}^b(\text{end}), \text{\textit{RobotPCD}$_a$}, \text{\textit{ScenePCD}})$ \\
        $\mathbf{q}^a \leftarrow \text{\textit{Add}}(\delta\mathbf{q}_{t} + \mathbf{q}^a(\text{end}))$ \\
        \textit{Complete} $\leftarrow$ \textit{Interpolation}($\mathbf{q}^a(\text{end}), \mathbf{q}^b(\text{end})$) \\
        \BlankLine
        \If{Complete}{
            $\mathbf{q}\leftarrow \text{\textit{Concatenate}}(\mathbf{q}^a, \mathbf{q}^b)$ \\
            \Return $\mathbf{q}$ \\
        }
        \Else{
            \Return $\emptyset$
        }
        \Comment*[l]{Bidirectional planning}
        swap($\mathbf{q}^a, \mathbf{q}^b$) 
    }
    \BlankLine
    \If{$\mathbf{q}$}{
        \Comment*[l]{Lazy path contraction}
        $\mathbf{q}\leftarrow$ \textit{PathContraction}($q$) \\
        \Comment*[l]{Path collision checking}
        \textit{CollisionFree} $\leftarrow$ \textit{CollisionChecking}($\mathbf{q}$) \\
        \BlankLine
        \If{CollisionFree}{
            \Return $\mathbf{q}$
        }\Else{
            \Comment*[l]{Replanning}
            $\mathbf{q}_{new}\leftarrow$ \textit{Replanning}($\mathbf{q}$) \\
            \Comment*[l]{Lazy path contraction}
            $\mathbf{q}_{new} \leftarrow$ \textit{PathContraction}($\mathbf{q}_{new}$) \\
            \Comment*[l]{Path collision checking}
            \textit{CollisionFree} $\leftarrow$ \textit{CollisionChecking}($\mathbf{q}_{new}$) \\
            \BlankLine
            \If{CollisionFree}{
                \Return $\mathbf{q}_{new}$
            }
        }
        \Return $\emptyset$
    }
\end{algorithm}%

\vspace{0.1cm}
\noindent
\textbf{Embodiment graph representation.} Similar to \cite{soleymanzadeh2025simpnet}, we utilize an undirected graph to model the kinematic structure of the robotic manipulator as demonstrated in Figure \ref{fig_2- graphconstruction}. We synthetically generate a manipulator point cloud by uniformly sampling on link meshes at arbitrary configurations, and apply the PointNet++ \cite{qi2017pointnet++} set abstraction layer to downsample the point cloud. In this representation, each point corresponds to a node, and edges are defined according to the manipulator's kinematic chain, such that each node is connected to its adjacent nodes along the kinematic chain of the manipulator. 

\vspace{0.1cm}
\noindent
\textbf{Spatial graph representation.} We utilize set abstraction layers from PointNet++ \cite{qi2017pointnet++} to downsample the workspace point cloud. We then consider each point with the downsampled workspace point cloud as a node, and construct a directed, fully connected graph that connects every workspace node to all the nodes of the manipulator, as demonstrated in Figure \ref{fig_2- graphconstruction}.

\subsection{\name Structure}
Given the adjacency matrix constructed in the previous section, we now describe each component of the proposed neural informed sampler.

\vspace{0.1cm}
\noindent
\textbf{Embedding robot and workspace planning information.} We utilize a shared multi-layer perceptron (MLP) to encode the current and planning goal configuration as follows:
\begin{equation} \label{eq - current}
\begin{aligned}
    \mathbf{z}_t = \text{MLP}_{I}(\mathbf{q}_t),
\end{aligned}
\end{equation}
\begin{equation} \label{eq - goal}
\begin{aligned}
    \mathbf{z}_{goal} = \text{MLP}_{I}(\mathbf{q}_{goal}),
\end{aligned}
\end{equation}
where $\mathbf{q}_t \in \mathbb{R}^6$, $\mathbf{q}_{goal} \in \mathbb{R}^6$, $\mathbf{z}_t \in \mathbb{R}^H$, and $\mathbf{z}_{goal} \in \mathbb{R}^H$ are current time-step configuration, goal configuration, current time-step configuration embedding, and goal configuration embedding, respectively. Also, we utilize set abstraction layer from PointNet++ \cite{qi2017pointnet++} to downsample and embed robot and workspace point clouds as follows:
\begin{equation} \label{eq - pcd_robot}
\begin{aligned}
    \mathbf{z}_r = \text{SetAbstraction}(\mathcal{P}_r),
\end{aligned}
\end{equation}
\begin{equation} \label{eq - pcd_scene}
\begin{aligned}
    \mathbf{z}_{w} = \text{SetAbstraction}(\mathcal{P}_w),
\end{aligned}
\end{equation}
where $\mathcal{P}_r \in \mathbb{R}^{N_r \times 3}$, $\mathcal{P}_w \in \mathbb{R}^{N_w \times 3}$, $\mathbf{z}_r \in \mathbb{R}^{K_r \times H}$, and $\mathbf{z}_w \in \mathbb{R}^{K_w \times H}$ are robot point cloud, scene point cloud, robot point cloud embedding, and scene point cloud embedding, respectively.

\vspace{0.1cm}
\noindent
\textbf{Transformer encoder.} After encoding all planning information, we calculate robot-related information as follows:
\begin{equation} \label{eq - robot}
\begin{aligned}
    \mathbf{z}_{robot} = \mathbf{z}_r + \mathbf{z}_t \otimes \mathbf{1}_{K_r} + \mathbf{z}_{goal} \otimes \mathbf{1}_{K_r},
\end{aligned}
\end{equation}
where $\mathbf{z}_{robot} \in \mathbb{R}^{K_r \times H}$ is robot embedding. Afterwards, we apply sinusoidal positional encoding to preserve the positional information of the transformer encoder input, since transformers are inherently agnostic to the spatial locations of their inputs \cite{dosovitskiy2020image}.

The vanilla self-attention mechanism in transformer architectures implicitly models a fully connected graph \cite{vaswani2017attention}. In this work, we bias the scaled dot-product attention to incorporate manipulator embodiment and planning spatial structure as follows:
\begin{equation} \label{eq - robot}
\begin{aligned}
    \text{Attention}(Q, K, V) = \text{softmax}(\frac{QK^T}{\sqrt{d_k}} + B)V,
\end{aligned}
\end{equation}
where $Q,~K$, and $V$ are query, key, and value matrices, respectively, with $d_k$ being the dimensionality of the key vector. The matrix $B$ is defined based on the adjacency matrix of the constructed graph from the previous section as follows:
\begin{equation} \label{eq - mask}
    B_{i,j} = 
    \begin{cases}
        0, & A_{i,j} = 1 \\
        -\infty, & A_{i,j} = 0
    \end{cases},
\end{equation}
which is equivalent to using the adjacency matrix ($A$) as a binary mask within the attention mechanism \cite{sferrazza2024body}. The transformer encoder interleaves layers with masked attention with layers with unmasked attention, starting with a masked attention layer as the first layer.

\vspace{0.1cm}
\noindent
\textbf{Transformer decoder.} The transformer decoder follows a standard transformer architecture and is conditioned on the transformer encoder output to predict motion planning actions. It takes as input a learnable token $\mathbf{s} \in \mathbb{R}^H$ and, using the encoder output as memory, generates a delta joint angle $\mathbf{\delta q}_t \in \mathbb{R}^6$ which is then converted into joint targets, and added to the constructed planning tree toward the goal.

\subsection{Stochasticity with Dropout}
We incorporate Dropout \cite{srivastava2014dropout} into the \name framework to induce random sample generation during deployment. This randomness is a defining characteristic of classical sampling-based motion planning algorithms and underlies their probabilistic completeness \cite{lavalle2006planning}. By introducing stochasticity, each planning attempt by \name may produce a different path between any given start and goal configuration.

\begin{figure}[htbp]
    \centering
    \includegraphics[width=\linewidth]{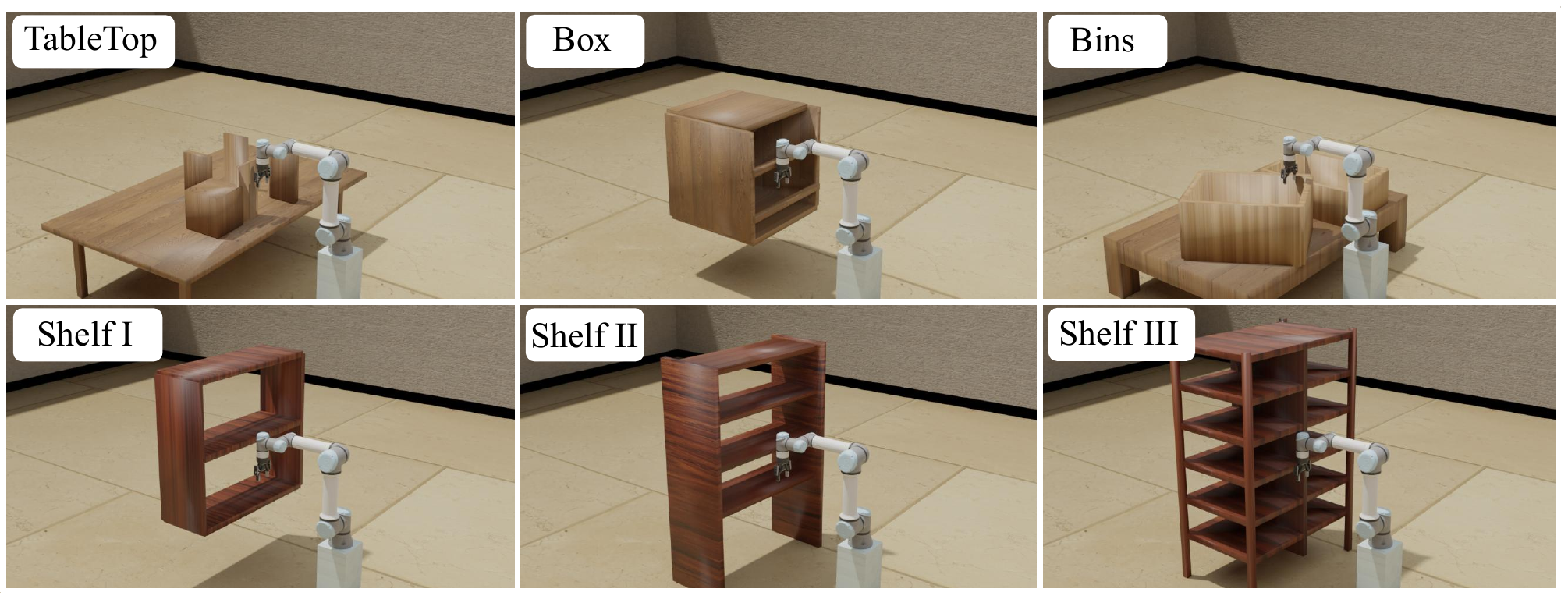}
    \caption{An example of all held-out planning tasks from \cite{soleymanzadeh2025perfact}.}
    \label{fig_4 - tasks}
    \vspace{-1em}
\end{figure}

\subsection{Bidirectional Planning Algorithm}
We embed the proposed neural sampler into the bidirectional planning algorithm proposed by \cite{qureshi2020motion} to plan between given start and goal configurations. Algorithm \ref{alg: simpt} outlines the overall planning algorithm. For a detailed description of the underlying bidirectional planning algorithm, we refer the reader to Qureshi \textit{et al.} \cite{qureshi2020motion}.
\section{Results and Discussion} \label{sec: discussions}
In this section, we describe the implementation details of \name and compare its performance with state-of-the-art sampling-based planning algorithms using uniform sampling, heuristic-based informed sampling, and neural informed sampling. All planning algorithms are implemented in PyTorch \cite{paszke2017automatic}, and the evaluations are conducted on a computer running Linux OS, equipped with an NVIDIA RTX 4080 GPU.

\setlength{\tabcolsep}{2pt}
\begin{table*}[htbp]
\begin{center}
\captionof{table}{Planning performance of \name and baseline planners across all held-out planning task environments.}

\label{tab: performance}

\hspace{-0.2cm}
\renewcommand{\arraystretch}{1.1}
\resizebox{\textwidth}{!}{%
\begin{tabular*}{\linewidth}{@{}l
@{\extracolsep{\fill}}
c c c c c c c@{\extracolsep{\fill}} c
c c c c c c}
\toprule
\phantom{Var.} &  
\multicolumn{2}{c}{\textbf{TableTop}}&\multicolumn{2}{c}{\textbf{Box}}&\multicolumn{2}{c}{\textbf{Bins}} && \multicolumn{6}{c}{\textbf{Shelf}}\\
\cmidrule{2-3}
\cmidrule{4-5}
\cmidrule{6-7}
\cmidrule{9-14}
\phantom{Var.}&\multicolumn{6}{c}{\phantom{Var.}}&&\multicolumn{2}{c}{\textbf{Task I}}&\multicolumn{2}{c}{\textbf{Task II}}&\multicolumn{2}{c}{\textbf{Task III}} \\
\cmidrule{9-14}
& {T $[s]\downarrow$} & {S $[\%]\uparrow$} & {T $[s]\downarrow$} & {S $[\%]\uparrow$} & {T $[s]\downarrow$} & {S $[\%]\uparrow$} & {} & {T $[s]\downarrow$} & {S $[\%]\uparrow$}& {T $[s]\downarrow$} & {S $[\%]\uparrow$}& {T $[s]\downarrow$} & {S $[\%]\uparrow$} \\
\toprule
\textbf{Bi-RRT} \cite{kuffner2000rrt}&$2.85 {\pm} 0.97$&$\mathbf{88}$\%&$\mathbf{0.73 {\pm} 0.54}$&$\mathbf{71}$\%&$\mathbf{1.1 {\pm} 0.67}$&$\mathbf{98.25}$\%&&$2.71 {\pm} 0.73$&$\mathbf{98}$\%&$3.91 {\pm} 1.12$&$\mathbf{82.6}$\%&$4.07 {\pm} 1.17$&$\mathbf{45}$\% \\
\textbf{RRT*} \cite{karaman2011sampling}&$4.15 {\pm} 0.99$&$50$\%&$2.59 {\pm} 0.07$&$55$\%&$3.08 {\pm} 0.1$&$64$\%&&$4.29 {\pm} 1.14$&$51$\%&$5.78 {\pm} 0.85$&$40.6$\%&$5.12 {\pm} 1.39$&$29$\% \\
\toprule
\textbf{IRRT*} \cite{gammell2014informed}&$4.44 {\pm} 1.35$&$48$\%&$2.54 {\pm} 0.8$&$55$\%&$3.12 {\pm} 0.09$&$66.5$\%&&$4.74 {\pm} 1.89$&$55$\%&$5.94 {\pm} 1.04$&$37$\%&$6.65 {\pm} 1.69$&$27$\% \\
\textbf{BIT*} \cite{gammell2020batch}&$3.19 {\pm} 2.67$&$71$\%&$1.4 {\pm} 1.47$&$67$\%&$1.84 {\pm} 1.81$&$93.75$\%&&$4.35 {\pm} 2.38$&$91$\%&$5.95 {\pm} 2.09$&$75$\%&$4.84 {\pm} 3.24$&$38$\% \\
\toprule
\textbf{MPNets} \cite{qureshi2020motion}&$\mathbf{2.66 {\pm} 1.26}$&$41$\%&$2.41 {\pm} 1.01$&$62$\%&$3.45 {\pm} 1.62$&$84.5$\%&&$\mathbf{2.29 {\pm} 0.01}$&$39$\%&$\mathbf{2.57 {\pm} 0.89}$&$34$\%&$\mathbf{3.43 {\pm} 1.19}$&$32$\% \\
\textbf{SIMPNet} \cite{soleymanzadeh2025simpnet}&$5.68 {\pm} 10.2$&$51$\%&$2.68 {\pm} 1.66$&$67$\%&$3.97 {\pm} 2.52$&$94.2$\%&&$3.28 {\pm} 0.1$&$44$\%&$2.96 {\pm} 1.43$&$35$\%&$6.88 {\pm} 4.78$&$33$\% \\
\toprule
\textbf{\name (ours)}&$3.0 {\pm} 2.58$&$52$\%&$2.17 {\pm} 2.02$&$65$\%&$3.72 {\pm} 3.56$&$96$\%&&$2.99 {\pm} 1.58$&$55$\%&$5.56 {\pm} 2.57$&$44$\%&$4.34 {\pm} 3.25$&$38$\% \\
\bottomrule
\end{tabular*}}
\end{center}
\end{table*}

\begin{figure*}
    \centering
    \includegraphics[width=0.9\linewidth]{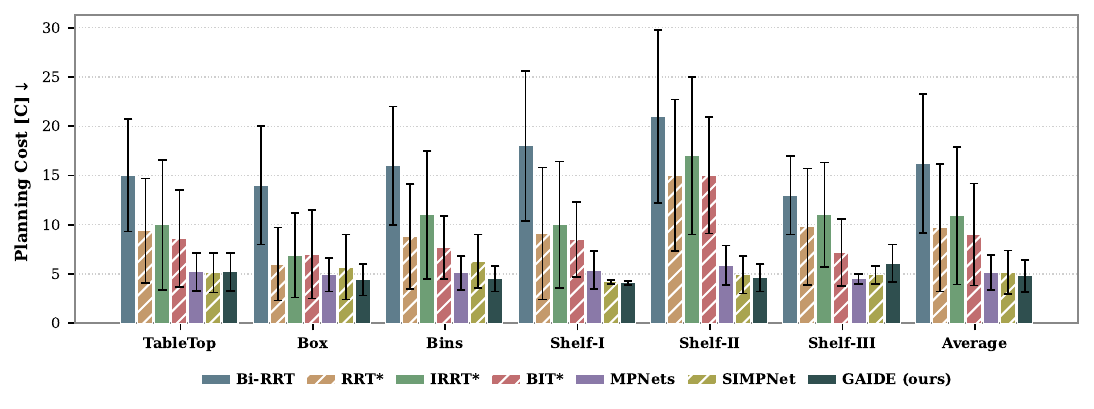}
    \caption{Planning cost of \name and baseline planners across all held-out planning tasks.}
    \label{fig_5 - planningcost}
    \vspace{-1em}
\end{figure*}

\subsection{Data Collection}
We utilize the scene generation framework from \cite{soleymanzadeh2025perfact} to construct a diverse set of motion planning workspaces and planning problems. We then employ cuRobo \cite{sundaralingam2023curobo}, a GPU-accelerated motion planner, to generate a dataset for training the proposed neural sampler.

\begin{figure*}[t]
    \centering
    \includegraphics[width=0.9\linewidth]{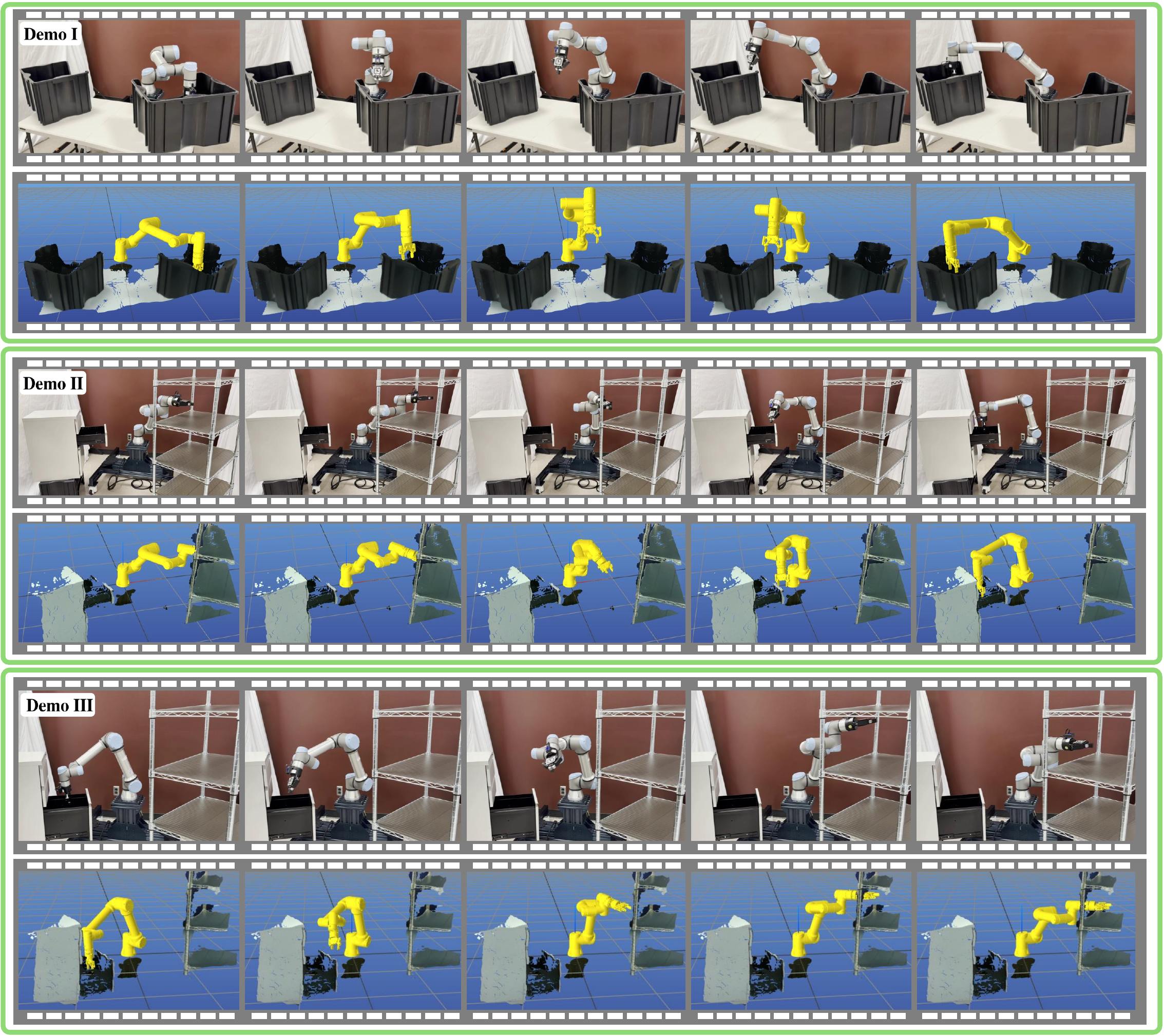}
    \caption{\textbf{Real-world deployment of \name}. The path profile from the start configuration to the goal configuration is demonstrated in the real-world (top), and simulated scene point cloud (bottom) for each demo.}
    \label{fig_7 - realdemo}
    \vspace{-1em}
\end{figure*}

\subsection{\name Training}
We utilize the \name parameterized by $\theta$ for informed sample generation within sampling-based planning algorithms. Let $\delta \mathbf{q}_{p,t} \in \mathbb{R}^6$ denote the ground-truth delta joint action at step $t$ along path $\mathbf{p}$ from the planning dataset $\mathcal{D} = \{\mathbf{p}_1, \cdots, \mathbf{p}_N\}$ where $\mathbf{p} = [\mathbf{q}_{p,0}, \cdots, \mathbf{q}_{p,t}, \cdots, \mathbf{q}_{p,T}]$. The loss function is defined as:
\begin{equation} \label{eq - lossfn}
\begin{aligned}
    \mathcal{L}_{\text{\name}} = \frac{1}{N_p}\sum_{p=1}^{N_p}\sum_{t=0}^{T_p - 1} \parallel \delta \mathbf{q}_{p, t} - \delta \hat{\mathbf{q}}_{p,t} \parallel^2,
\end{aligned}
\end{equation}
where $N_p$ is the batch size. The model is optimized using standard MSE loss for approximately 1M gradient steps, which requires wall-clock time of one day training on a single NVIDIA A100 GPU with a batch size of 256.

\subsection{Baselines and Metrics}
\noindent
\textbf{Baselines.} We evaluate the performance of \name by comparing it with several state-of-the-art sampling-based motion planning algorithms. For planners using uniform sampling, we consider Bidirectional RRT (Bi-RRT) \cite{kuffner2000rrt} and RRT* \cite{karaman2011sampling}. For heuristic-based informed sampling, we include Informed RRT* (IRRT*) \cite{gammell2014informed} and Batch Informed Trees (BIT*) \cite{gammell2020batch}. All classical sampling-based algorithms are implemented using Open Motion Planning Library (OMPL) \cite{sucan2012open}. Since these planners lack internal termination conditions, we set the planning time for these planners to match the average planning time of \name for each planning task. For neural informed sampling, we evaluate against MPNets \cite{qureshi2020motion}, and SIMPNet \cite{soleymanzadeh2025simpnet}. These benchmark planners are selected to highlight the effectiveness of incorporating spatial structure and the manipulator's kinematic chain into neural-informed sampling framework. All neural informed samplers are embedded within the same bi-directional planner \cite{qureshi2020motion}, and use the same workspace embedding network. All planners utilize PyBullet \cite{coumans2016pybullet} physics engine for collision checking.

\vspace{0.1cm}
\noindent
\textbf{Metrics.} We evaluate \name using three standard planning metrics: \textit{planning time}, \textit{planning cost} and \textit{success rate}. \textit{planning time} ``T'' denotes the average planning time the planner takes in each evaluation task. \textit{Planning cost} ``C'' measures the length of the successfully planned paths within the configuration space. \textit{Success rate} ``S'' represents the percentage of successfully planned paths.

\subsection{Evaluation Results}
\begin{figure}
    \centering
    \includegraphics[width=\linewidth]{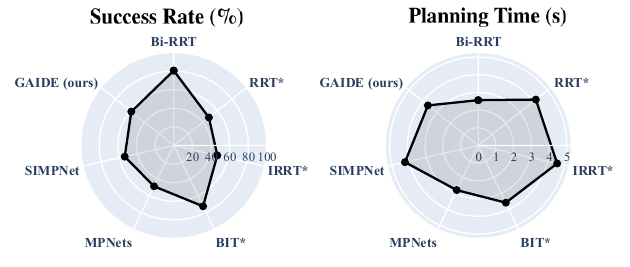}
    \caption{Average success rate and planning time comparison between \name and benchmark planners across all held-out planning tasks.}
    \label{fig_6 - average}
    \vspace{-2em}
\end{figure}

We utilize held-out planning environments proposed by \cite{soleymanzadeh2025perfact} to evaluate the performance of \name against benchmark planners. Figure \ref{fig_4 - tasks} illustrate an example of these planning tasks. Table \ref{tab: performance} and Figures \ref{fig_5 - planningcost} and \ref{fig_6 - average} report the performance of \name in comparison with benchmark motion planners across these planning tasks.

\vspace{0.1cm}
\noindent
\textbf{\name vs. uniform samplers.} Bi-RRT achieves the fastest planning time and the highest success rate across all planning tasks due to its bidirectional tree construction module. However, Bi-RRT terminates once an initial feasible solution is found, which results in suboptimal solutions and higher planning costs. Compared to \name, Bi-RRT attains higher success rates and lower planning times as demonstrated in Table \ref{tab: performance}. However, it consistently exhibits the worst planning cost among benchmark planners (on average, Bi-RRT: \underline{$16.2 {\pm} 7.05$} vs. \name: \underline{$4.81 {\pm} 1.63$}).

RRT* is an asymptotically optimal planner that utilizes uniform sampling and graph rewiring to improve solution quality within the given planning time budget. However, both uniform sampling and graph rewiring modules are computationally expensive, which results in the lowest success rate among virtually all benchmark planner across all planning tasks, as demonstrated in Table \ref{tab: performance}. Moreover, under the planning time budget, RRT* yields higher planning cost than \name across all evaluation tasks (on average RRT*: \underline{$6.98 {\pm} 6.47$} vs. \name: \underline{$4.81 {\pm} 1.63$}).

\vspace{0.1cm}
\noindent
\textbf{\name vs. heuristic-based informed samplers.} Informed RRT* (RRT*) is also an asymptotically optimal planner that combines informed sampling with graph rewiring to improve solution quality within the given planning time budget. Although IRRT* restricts the sampling to an informed subset after an initial feasible solution is found, the rewiring module remains computationally expensive. As a result IRRT* exhibits the lowest success rate among virtually all benchmark planner across all planning tasks, as demonstrated in Table \ref{tab: performance}. Moreover, under the planning time budget, IRRT* yields higher planning cost than \name across all evaluation tasks (on average IRRT*: \underline{$10.9 {\pm} 6.98$} vs. \name: \underline{$4.81 {\pm} 1.63$}).

BIT* is an asymptotically optimal planner that performs batch-wise, heuristic-guided search (informed) over a random geometric graph, which leads to high success rates and efficient planning times across all planning tasks, as demonstrated in Table \ref{tab: performance}. However, under the given planning budget, BIT* yields higher planning cost than \name across all evaluation tasks (on average: BIT*: \underline{$9.0 {\pm} 5.17$} vs. \name: \underline{$4.81 {\pm} 1.63$})

\vspace{0.1cm}
\noindent
\textbf{\name vs. neural informed samplers}: MPNets simply concatenates planning-related information as input to an MLP-based neural informed sampling, without explicitly encoding the spatial and kinematic structure inherent within the motion planning problems. As a result, MPNets exhibits lower success rate compared to \name across all planning tasks, as demonstrated in Table \ref{tab: performance}. SIMPNet, on the other hand, encodes the kinematic structure of the robotic manipulator to establish kinematic-aware sampling. However, GNNs' struggle with encoding long-horizon interactions due to representation oversmoothing or overquashing leads to lower success rates compared to \name across all planning tasks, as demonstrated in Table \ref{tab: performance}.

\subsection{Ablation Study}
In this section, we compare \name with its ablated variants to evaluate the contribution of attention masking to the performance of the neural sampler. These ablation frameworks are as follows:

\vspace{0.1cm}
\noindent
\textbf{\name-Vanilla Transformer.} Inspired by \cite{zhao2023learning}, this framework features a standard encoder-decoder transformer architecture for predicting future informed samples given planning information. It processes current-time step planning information using a vanilla attention mechanism and does not incorporate the spatial and kinematic structure of the planning problem.

\vspace{0.1cm}
\noindent
\textbf{\name-Hard.} This framework utilizes the same attention mask as \name, but applies it at every transformer encoder layer, instead of interleaving masked and standard attention layers.

As shown in Table~\ref{tab: ablations}, \name-V achieves lower success rates compared to \name across all evaluated planning tasks. This performance drop is due to the fact that motion planning problem depends on the inherent spatial structure of the problem, and explicitly incorporating such spatial information into the motion planning framework improves planning efficiency. 

However, incorporating the attention mask at every layer of the transformer encoder (i.e., \name-H) leads to a deteriorated planning performance across all planning tasks, even compared to the ablation variant without attention masking (i.e., \name-V), as demonstrated in Table~\ref{tab: ablations}. This performance drop in terms of success rates can be attributed to the fact that the constructed attention mask restricts the transformer decoder's ability to fully attend to the workspace information embedding, causing some spatial information to be masked out at each encoder layer.

\setlength{\tabcolsep}{2pt}
\begin{table*}[htbp]
\begin{center}
\captionof{table}{Planning performance of \name and different frameworks that apply attention masking differently across held-out planning environments. ``\name-V'' denotes \name-Vanilla Transformer, and ``\name-H'' denotes \name-Hard.}

\label{tab: ablations}

\hspace{-0.2cm}
\renewcommand{\arraystretch}{1.1}
\resizebox{\textwidth}{!}{%
\begin{tabular*}{\linewidth}{@{}l
@{\extracolsep{\fill}}
c c c c c c c@{\extracolsep{\fill}} c
c c c c c c}
\toprule
\phantom{Var.} &  
\multicolumn{2}{c}{\textbf{TableTop}}&\multicolumn{2}{c}{\textbf{Box}}&\multicolumn{2}{c}{\textbf{Bins}} && \multicolumn{6}{c}{\textbf{Shelf}}\\
\cmidrule{2-3}
\cmidrule{4-5}
\cmidrule{6-7}
\cmidrule{9-14}
\phantom{Var.}&\multicolumn{6}{c}{\phantom{Var.}}&&\multicolumn{2}{c}{\textbf{Task I}}&\multicolumn{2}{c}{\textbf{Task II}}&\multicolumn{2}{c}{\textbf{Task III}} \\
\cmidrule{9-14}
& {T $[s]\downarrow$} & {S $[\%]\uparrow$} & {T $[s]\downarrow$} & {S $[\%]\uparrow$} & {T $[s]\downarrow$} & {S $[\%]\uparrow$} & {} & {T $[s]\downarrow$} & {S $[\%]\uparrow$}& {T $[s]\downarrow$} & {S $[\%]\uparrow$}& {T $[s]\downarrow$} & {S $[\%]\uparrow$} \\
\toprule
\textbf{\name-V}&$2.96 {\pm} 2.71$&$45$\%&$2.15 {\pm} 1.51$&$62$\%&$3.61 {\pm} 3.58$&$81.25$\%&&$3.26 {\pm} 1.56$&$45$\%&$\mathbf{3.58 {\pm} 2.79}$&$36$\%&$\mathbf{4.03 {\pm} 3.42}$&$33$\% \\
\toprule
\textbf{\name-H}&$\mathbf{2.61 {\pm} 2.71}$&$48$\%&$\mathbf{2.01 {\pm} 1.25}$&$59$\%&$\mathbf{3.41 {\pm} 3.04}$&$77.25$\%&&$\mathbf{2.48 {\pm} 1.51}$&$45$\%&$4.58 {\pm} 2.29$&$34$\%&$4.42 {\pm} 4.98$&$30$\% \\
\toprule
\textbf{\name}&$3.0 {\pm} 2.58$&$\mathbf{52}$\%&$2.17 {\pm} 2.02$&$\mathbf{65}$\%&$3.72 {\pm} 3.56$&$\mathbf{96}$\%&&$2.99 {\pm} 1.58$&$\mathbf{55}$\%&$5.56 {\pm} 2.57$&$\mathbf{44}$\%&$4.34 {\pm} 3.25$&$\mathbf{38}$\% \\
\bottomrule
\end{tabular*}}
\end{center}
\end{table*}

\subsection{Real-world Deployment}
To evaluate the performance of our planner on a physical robot, we conducted experiments in a real-world environment (Figure \ref{fig_7 - realdemo}). The scene was represented using point cloud data acquired from a calibrated Intel RealSense D435i RGB-D camera. Collision checking was performed using a spherical approximation of the robot geometry. The transformation between the camera and the robot base was estimated using AprilTag markers. The results demonstrate that \name generalizes effectively to real-world sensor data without additional training or fine-tuning.
\section{Conclusions} \label{sec: conclusions}
In this paper, we presented \name, a neural informed sampler that can be embedded within sampling-based motion planners for efficient motion planning. We constructed a graph that represents the manipulator's kinematic chain and the spatial structure inherent within motion planning problems, and deployed its adjacency matrix as an attention mask to incorporate these inherent structures into the neural sampler. \name was trained on optimal paths generated by an oracle planner via supervised learning, and employed dropout during inference to introduce stochasticity into the planning algorithm.

We evaluated \name by embedding it into a bidirectional motion planner and comparing its performance with state-of-the-art sampling-based motion planners using uniform sampling, heuristic-based informed sampling, and neural informed sampling. The results demonstrate that \name achieves superior performance compared to benchmark motion planners across all held-out evaluation tasks. Ablation studies further demonstrate that explicitly incorporating the spatial and kinematic structure of the motion planning problem into the neural sampler leads to substantial performance improvement.
\bibliographystyle{IEEEtran}
\bibliography{ref}

\end{document}